# Implementation of Rule Based Algorithm for Sandhi-Vicheda Of Compound Hindi Words

Priyanka Gupta[1] ,Vishal Goyal [2]
[1]*M.Tech. (ICT) Student,* [2]*Lecturer*
*Department of Computer Science*
*Punjabi University Patiala*
[1]*priyankaquick@gmail.com ,* [2]*vishal.pup@gmail.com*

## Abstract

Sandhi means to join two or more words to coin new word. Sandhi literally means `putting together' or combining (of sounds), It denotes all combinatory sound-changes effected (spontaneously) for ease of pronunciation. Sandhi-vicheda describes [5] the process by which one letter (whether single or cojoined) is broken to form two words. Part of the broken letter remains as the last letter of the first word and part of the letter forms the first letter of the next letter. Sandhi-Vicheda is an easy and interesting way that can give entirely new dimension that add new way to traditional approach to Hindi Teaching. In this paper using the Rule based algorithm we have reported an accuracy of 60-80% depending upon the number of rules to be implemented.

## I INTRODUCTION

Natural Language Processing (NLP) refers to descriptions that attempt to make the computers analyze, understand and generate natural languages, enabling one to address a computer in a manner as one is addressing a human being. Natural Language Processing is both a modern computational technology and a method of investigating and evaluating claims about human language itself. It is a subfield of artificial intelligence and computational linguistics. It studies the problems of automated generation and understanding of natural human languages.

A word can be defined as a sequence of characters delimited by spaces, punctuation marks, etc. in case of written text. A compound word (also known as co-joined word) can be broken up into two or more independent words. A Sandhi-Vicheda module breaks the compound word in a sentence into constituent words. Sandhis take place whenever there is a presence of a swara i.e.a vowel; the presence of a consonant with a halanta; the presence of a visarga. Sanskrit has a well defined set of rules for Sandhi-vicheda. But Hindi has its own rules of Sandhi-vicheda. They are, however, not so well-defined as, and much fewer in number than, the Sanskrit rules.

### 1.1 The Hindi Language

Hindi is spoken in northern and central India. Linguists think of Hindi and Urdu as the same language, the difference being that Hindi [5] is written in the Devanagari script and draws much of its vocabulary from Sanskrit, while Urdu is written in the Persian script and draws a great deal of its vocabulary from Persian and Arabic. More than 180 million people in India regard Hindi as their mother tongue. Another 300 million use it as second language. Hindi is the national language of India and is spoken by almost half a billion people in India and throughout the world and is the world's second most spoken language. It allows you to communicate with a far wider variety of people in India than English which is only spoken by around five percent of the population. It is written in an easy to learn phonetic script called "*Devanagari*" which is also used to write Sanskrit, Marathi and Nepali. Hindi is normally spoken using a combination of 52 sounds, ten vowels, 40 consonants, nasalisation and a kind of aspiration. These sounds are represented in the Devanagari script by 52 symbols: for ten vowels, two modifiers and 40 consonants.

## II RELATED WORK

**Sandhi** (in linguistics) [1] is a cover term for a wide variety of phonological processes that occur at morpheme or word boundaries, such as the fusion of sounds across word boundaries and the alteration of sounds due to neighboring sounds or due to the grammatical function of adjacent words. **Internal sandhi** features the alteration of sounds within words at morpheme boundaries, as in *sympathy* (*syn-* + *pathy*). **External sandhi** refers to changes found at word boundaries, such as in the pronunciation [tɛm bʊks] for *ten books*. This is not true of all dialects of English. The *Linking R* of some dialects of English is a kind of external sandhi, as is the process called *liaison* in the French language. While it may be extremely common in speech, sandhi (especially external) is typically ignored in spelling, as is the case in English, with the exception of the distinction between "a" and "an" (sandhi is, however, reflected in the writing system of Sanskrit and Hindi). External sandhi effects can sometimes become morphologized. Most tonal languages have **Tone sandhi**, in which the tones of words alter according to pre-determined rules. For example: Mandarin has four tones: a high monotone, a rising tone, a falling-rising tone, and a falling tone. In the common greeting *nĭ hăo*, both words in isolation would normally have the falling-rising tone. However, this is difficult to say, so the tone on *nĭ* is pronounced as *ní* (but still written nĭ in <u>Hanyu Pinyin</u>).

**The Sanskrit Sandhi engine software** is not currently available as a standalone application, since its local use demands the installation of an HTTP server on the user's host.

**The Sandhi module**[1] developed by RCILTS-Sanskrit, Japanese, Chinese at Jawaharlal Nehru University, New Delhi. RCILTS, JNU is a resource center for Sanskrit language of DIT, Government of India. At JNU work started in three languages viz., Sanskrit, Japanese, and Chinese. Using this module the user can get the information about *Sandhi* rules and processes. *Sutra* number in *Astyadhayi* and its description is displayed. User can learn three types of Svara Sandhi, Vyanjan Sandhi, Hal Sandhi through this *Sandhi* module Data is in Unicode. *Sandhi* exceptions and options are also incorporated. This module takes two words as input. First word cannot be null but second word can be. A user can input the two words and submit the form to get the result of the given input.

**Chinese Tone Sandhi**,[2] Cheng and Chin-Chuan from California University, Berkeley, Phonology Laboratory faced the problem that English stresses are interpreted by Chinese speakers when they speak Chinese with Engish words inserted. Chinese speakers in the United States usually speak Chinese with Engish words inserted. In Mandarin Chinese, a tone-sandhi rule changes a third tone preceding another third tone to a second tone. Using the tone-sandhi rule, they designed the experiment to find out hoe English stresses are interpreted in Chinese sentences. Stress does not exist in the underlying representations of English phonology. But in studying bilingual phenomena, the phonetic level is also important. Fry (1995) found that when a vowel was long and of high intensity, listeners agreed that the vowel was strongly stressed. The results of his experiments indicate that the duration ratio has a stronger influence on judgements of stress than has the intensity ratio. Lehiste and Peterson (1959) also reported experiments on stress.

**English l-sandhi** [3] involves an allophonic alternation in alveolar contact for word-final /l/ in connected speech [4]. EPG data for five Scottish Standard English and five Southern Standard British English speakers shows that there is individual and dialectal variation in contact patterns.

## III PROBLEM DEFINITION

Developing programs that understand a natural language is a difficult task. Natural languages are large. They contain an infinity of different sentences. No matter how many sentences a person has heard or seen, new ones can always be produced. Also, there is much ambiguity in a natural language. Many words have several meanings and sentences can have different meanings in different contexts. Compound words are created by joining an arbitrary number of existing words together, and this can lead to a large increase of the vocabulary size, and thus also to sparse data problems. Therefore the problem of compound words poses challenges for many NLP applications. The problem domain, to which this paper is concerned, is breaking up of Hindi compound words into constituent words. In Hindi, words are a sequence of characters. These words are combined with 'swar', 'vyanjan', and matra's. Hindi has its own rules of Sandhi-vicheda. They are, however, not so well-defined as, and much fewer in number than, the Sanskrit rules. So my problem is to break the compound word into constituent words with the help of rules of 'Sandhi-vicheda' in Hindi grammar. My problem is to design a Graphical User Interface, which accepts input as a Hindi language word (source text) from the keyboard or mouse and break it into constituent words (target text). The source text is converted into target text in **Unicode Format**.

| Compound Word | Sandhi-vicheda |
|---|---|
| ijk/khu | ij $ v/khu |
| HkkokFkZ | Hkko $ vFkZ |
| f'koky; | f'ko $ vky; |
| dohUnz | dfo $ bUnz |
| x.ks'k | x.k $ bZ'k |
| ijes'oj | ije $ bZ'oj |
| ,dSd | ,d $ ,d |
| ;FkSd | ;Fkk $ ,d |
| ijksidkj | ij $ midkj |
| lfU/kPNsn | lfU/k $ Nsn |

| foPNsn | fo $ Nsn |
|--------|----------|

Table 1:Sandhi-Vicheda of Hindi Compound Words

## IV IMPLEMENTATION

We have implemented the *Rule-Based algorithm* to first manually find the compound words and then develop the program that uses the database for displaying the correct meaning to the Sandhi-Vicheda word according to the Hindi grammar Sandhi-Vicheda rules.

### 4.1 Algorithm

hword = hindi word to be entered
cur = Variable that stores the length of string

**Step 1:** Repeat for every word of the input string.
**Step 2:** Count the Length of String.
**Step 2.1:** Store the Length of String in variable.
       For i = 1 To Len(hword)
       cur = Mid$(hword, i, 1)

**Step 3:** Find the position of Matra.

       hword.Substring(b - 1, 1)
**Step 4:** Apply the rules for sandhi –vicheda

**Step 4.1:** (Rule for "Sign-AA( **k** )" replaced with Swar "Letter-A( **v** )" in Sandhi vicheda)

| LokFkhZ | Lo $ vFkhZ |
|---------|-----------|
| HkkokFk Z | Hkko $ vFkZ |
| IR;kFkhZ | IR; $ vFkhZ |
| ;FkkFkZ | ;Fkk $ vFkZ |

Table 2: Rule I Implemented Word List

**Step 4.2:** (Rule for "Sign-AA( **k** )" replaced with Swar "Letter-AA( **vk** )" in Sandhi vicheda)

| fo|ky; | fo|k $ vky; |
|--------|-----------|
| f'koky; | f'ko $ vky; |

| iqLrdky; | iqLrd $ vky; |
|----------|-------------|
| Hkkkstuky; | Hkkkstu $ vky; |

Table 3: Rule II Implemented Word List

**Step 4.3:** (Rule for "Sign-E( **h** )" replaced with Swar "Letter-E( **b** )" in Sandhi vicheda)

| ujsUnz | uj $ bUnz |
|--------|-----------|
| lqjsUnz | lqj $ bUnz |
| dohUnz | dfo $ bUnz |
| 'kphUnz | 'kph $ bUnz |

Table 4: Rule III Implemented Word List

**Step 4.4:** (Rule for "Sign-E( **h** )" replaced with Swar "Letter-E( ई )" in Sandhi vicheda)

| fxjh'k | fxfj $ bZ'k |
|--------|-----------|
| jtuh'k | jtuh $ bZ'k |
| x.ks'k | x.k $ bZ'k |
| ijes'oj | ije $ bZ'oj |

Table 5: Rule IV Implemented Word List

**Step 4.5:** (Rule for "Sign-U( **ks** )" replaced with "Letter-U( **m** )" in Sandhi Vicheda)

| ijksidkj | ij $ midkj |
|----------|-----------|
| egksnf/k | egk $ mnf/k |
| vkRekskRlxZ | vkRe $ mRlxZ |
| lkxjkskseZ | lkxj $ meZ |

Table 6: Rule V Implemented Word List

**Step 4.6:** (Rule for "Sign-EE( **S** )" replaced with Vowel "Letter-E( **,** )" in Sandhi Vicheda)

| lnSo | lnk $ ,o |
|------|----------|

| egSo | egk $ ,o |
|------|----------|
| ;FkSo | ;Fkk $ ,o |
| ,dSd | ,d $ ,d |

Table 7: Rule VI Implemented Word List

**Step 4.7:** (Rule for "Sign-EE ( **S** )" replaced with "Letter-EE ( **,s** )" in Sandhi Vicheda)

| egS'o;Z | egk $ ,s'o;Z |
|---------|--------------|
| nsoS'o;Z | nso $ ,s'o;Z |
| ijeS'o;Z | ije $ ,s'o;Z |
| ;FkSfrgkfld | ;Fkk$ ,sfrgkfld |

Table 8: Rule VII Implemented Word List

**Step 4.8:** (Rule for eliminating the half letter in Sandhi- Vicheda) If find the (Half CH) (**P**) Letter then eliminates the Letter and decompose the word.

| lfU/kPNsn | lfU/k $ Nsn |
|-----------|-------------|
| foPNsn | fo $ Nsn |
| ifjPNsn | ifj $ Nsn |
| y{ehPNk;k | y{eh $ Nk;k |

Table 9: Rule VIII Implemented Word List

**Step 4.9:** (Rule of Visarga in Sandhi Vicheda) If find the (Half Letter) then replace with Sign ( **:** )visarga.

| fu'py | fu% $ py |
|-------|----------|
| fu'rst | fu% $ rst |
| nqLlkgl | nq% $ lkgl |
| fuLrkj | fu% $ rkj |

Table 10: Rule IX Implemented Word List

**Step 5:** Repeat Steps 4.1 to 4.9 to check the next word for checking the Vyanjan that combined with Matra. Then replace the Matra with Swar.

**Step 6:** Find the Unicode value for each of the Hindi characters and additional characters and use those values to implement above rules.

**Step 7:** Display the results.

Our module was developed in Visual Basic.NET (2005) and the encoding used for text was in Unicode, most suitable for other applications as well. Unicode uses a 16 bit encoding that provision for 65536 characters. Unicode standard [18] assigns each character a unique numeric value and name. Presently it provides codes for 49194 characters:

In Hindi Language:  Total Swar=13
Total Vyanjan=33
Total Matra=13

## V RESULTS AND DISCUSSION

We have tested our software on more than 200 words. Using the Rule based algorithm we have reported an accuracy of 60-80% depending upon the number of rules to be implemented. SANDHI-VICHEDA is an easy and interesting way that can give entirely new dimension that add new way to traditional approach to Hindi Teaching.

## VI CONCLUSION AND FUTURE WORK

In this paper, we presented the technique for the Sandhi-Vicheda of compound hindi words. Using the Rule based algorithm we have reported an accuracy of 60-80% depending upon the number of rules to be implemented. As future work, database can be extended to include more entries to improve the accuracy. This software can be used as a teaching aid to all the students from Class-V to the highest level of education. With this software one can learn about the very important aspect of Hindi Grammar i.e. 'SANDHI-VICHEDA'. By adding new more features, we can upgrade it to learn all the aspects of Hindi Grammar. It can also be used to solve and test the problems related to Hindi Grammar.

## ACKNOWLEDGEMENT

We would like to thank Dr. G.S. Lehal, Professor and Head, Department of Computer Science, Punjabi University, Patiala for many helpful suggestions and comments.

# REFERENCES

[1] Bharati, Akshar, Vineet Chaitanya & Rajeev Sangal, 1991, *A Computational Grammar for Indian languages processing*, Indian Linguistics Journal, pp.52, 91-103.

[2] Bharati A., Chaitanya V and Sangal R, "Natural Language processing: A Paninian Perspective", Prentice Hall of India, 1995.

[3] Cheng, Chin-Chuan "English Stresses and Chinese Tones in Chinese Sentences" California University, Berkeley, Phonology Laboratory.

[4] Dan W. Patterson "Introduction to Artificial Intelligence and Expert Systems" Prentice Hall P-227.

[5] Elaine Rich, Kevin Knight "Artificial Intelligence" Tata McGraw-Hill Second Edition, P-377.

[6] Jain Vinish 2004, *Sanskrit-English Anus¡raka:Morphological Analyzer and Dictionary Component,IIIT-Hyderabad.*

[7] James M. Scobbie (Queen Margaret University), Marianne Pouplier (Edinburgh University), Alan A. Wrench (Articulate Instruments Ltd.) "Conditioning Factors in External Sandhi: An EPG Study of English /l/ Vocalisation".

[8] Jha, Girish N., 2004, *The system of Pa̅ini*, Language in India, volume4:2.

[9] Jha, Girish N. et al., 2006, *Towards a Computational analysis system for Sanskrit*, Proc. of *first National symposium on Modeling and Shallow parsing of Indian Languages* at Indian Institute of Technology Bombay, pp 25-34.

[10] Jurafsky Daniel and James H. Martin, 2000, *Speech and Languages Processing,* Prentice-Hall, New Delhi.

[11] Kasturi Venkateswara Rao, "A Web-Based Simple Sentence
Level GB Translator from Hindi to Sanskrit", M.Tech(CS) Dissertation, School of Computer Systems Sciences, Jawaharlal Nehru University, New Delhi.

[12] Mitkov Ruslan, *The Oxford Handbook of Computational Linguistics*, Oxford University Press.

[13] Peng, Shu-hui (1994). 'Effects of prosodic position and tonal context on Taiwanese Tones'. Ohio State University Working Papers in Linguistics, 44, 166-190.

[14] Resource Centre For Indian Language Technology Solutions Sanskrit, Japanese, Chinese Jawaharlal Nehru University, New Delhi "Achievements".

[15] Scobbie, J. & Wrench, A., 2003. "An articulatory investigation of word-final /l/ and /l/-sandhi in three dialects of English". Proc. XVth ICPhS, 1871-1874.

[16] Suraj Bhan Singh. Hindi bhasha: Sandharbh aur Sanrachna. Sahitya Sahakar,1991.

[17] Whitney, W.D., 2002, *History of Sanskrit Grammar*, Sanjay Prakashan, Delhi.